\documentclass[conference]{IEEEtran}
\usepackage{cite}
\usepackage{amsmath,amssymb}
\usepackage{tikz}
\usetikzlibrary{fit}
\usepackage{caption}
\usepackage{url}
\usepackage{hyperref}
\usepackage[hyphenbreaks]{breakurl}
\usetikzlibrary{positioning}
\usepackage{pgfplots}
\pgfplotsset{compat=1.9}

\begin{document}
\title{Attention in SRAM on Tenstorrent Grayskull}

\author{\IEEEauthorblockN{Moritz Thüning}
Technical University of Munich \\
moritz.thuening@tum.de}

\date{\today}

\maketitle

\begin{abstract}
\boldmath
When implementations of the Transformer's self-attention layer utilize SRAM instead of DRAM, they can achieve significant speedups. The Tenstorrent Grayskull architecture provides a large SRAM, distributed across a grid of cores. This work presents a fused kernel for Grayskull, that exclusively utilizes its large SRAM by combining matrix multiplication, attention score scaling and Softmax operations. Additionally, a dedicated Softmax kernel utilizing the SRAM and a CPU implementation serving as a baseline are presented. The Softmax operation consumes most of the runtime in the computation of attention weights from queries and keys on Grayskull. The speedup of the dedicated Softmax kernel compared to the CPU implementation is up to \(10 \times\), and the Softmax implementation inside the fused kernel is approximately \(1.8 \times\) faster than the dedicated Softmax kernel. The time and memory complexity of all implementations is quadratic in sequence length. Currently, the Grayskull e150 is approximately \(30 \times\) cheaper for the general public than an Nvidia H100 PCIe (a state-of-the-art GPU) and offers approximately \(1.5 \times\) more SRAM.
\end{abstract}

\section{Introduction}
The Transformer \cite{vaswani2017} has become the state-of-the-art architecture in many applications, particularly in natural language processing. However, it is based on the self-attention layer which has a time and memory complexity quadratic in sequence length.

To improve the complexity, approximate attention mechanisms have been proposed, but they are not efficient or accurate enough to be widely adopted \cite{kitaev2020, roy2021, choromanski2021, katharopoulos2020, wang2020, beltagy2020, chen2021, zaheer2020}. Another approach is to improve the memory bandwidth and latency without changing the quadratic complexity. The observation that the Softmax operation in self-attention is memory-bound on GPUs led to FlashAttention \cite{dao2022} which achieved significant speedups. It is an exact attention algorithm that uses tiling and in the backward pass recomputation to reduce the movement of attention scores between the GPU's high-bandwidth memory (HBM) and SRAM.

The recent increase in demand for AI applications motivated the design of new hardware architectures to accelerate highly parallel AI workloads. One of those is the Tenstorrent Grayskull architecture \cite{vasiljevic2021, tenstorrent-cards, grayskull-architecture, grayskull-noc}, commercially available for example as a Grayskull e150 PCIe card. Currently, it is approximately \(30\times\) cheaper than the Nvidia H100 PCIe card \cite{nvidia2022} (a state-of-the-art GPU). Nevertheless, the Grayskull e150 provides \(120\,\text{MB}\) SRAM (120 Tensix cores with \(1\,\text{MB}\) each) compared to only \(80\,\text{MB}\) of the Nvidia H100 PCIe (114 Streaming Multiprocessors with \(256\,\text{KB}\) L1 each and \(50\,\text{MB}\) L2 shared). 

This background provided the motivation for this work, which includes the implementation of a \textbf{dedicated Softmax kernel} leveraging the large SRAM of the Grayskull e150, as well as a \textbf{fused kernel} combining matrix multiplication, attention score scaling and Softmax operations. The fused kernel reduces overhead of dispatching kernels and of moving inputs/outputs from separate kernels to and from DRAM.

The code is available at \url{https://github.com/moritztng/grayskull-attention}.

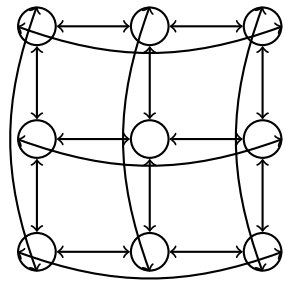
\begin{figure}[h]
\centering
\begin{tikzpicture}[
    scale=0.5,
    every node/.style={circle, draw, minimum size=0.5cm},
    bend angle=20,
    thick,
]
\foreach \row in {0,1,2} {
    \foreach \col in {0,1,2} {
        \node (\row-\col) at (3*\col, -3*\row) {};
    }
}

\foreach \row in {0,1,2} {
    \foreach \col in {0,1,2} {
        \ifnum\col<2
            \draw[<->] (\row-\col.east) -- (\row-\the\numexpr\col+1\relax.west);
        \else
            \draw[<->, bend left] (\row-\col.east) to (\row-0.west);
        \fi
        \ifnum\row<2
            \draw[<->] (\row-\col.south) -- (\the\numexpr\row+1\relax-\col.north);
        \else
            \draw[<->, bend left] (\row-\col.south) to (0-\col.north);
        \fi
    }
}
\end{tikzpicture}
\caption{Topology of the Network-on-Chip (NoC). Nodes represent Tensix cores and the edges represent bi-directional connections between them. The actual Tensix core grid of Grayskull is \(10 \times 12\). It is a torus topology, since opposite ends are connected.}
\label{fig:noc_topology}
\end{figure}

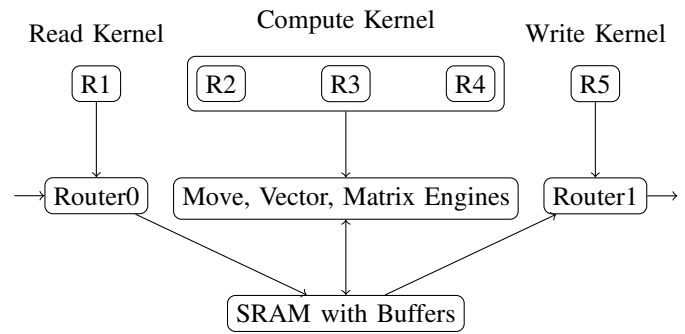
\begin{figure}[h]
\centering
\begin{tikzpicture}[
    node distance=1cm,
    smallnode/.style={
        draw,
        rounded corners=3pt,
        text centered,
    },
    container/.style={
        draw,
        rounded corners=3pt,
        fit={(r2) (r3) (r4)},
    },
    kernel/.style={
        anchor=south,
        above=5pt
    }
]

\node[smallnode] (r1) {R1};
\node[smallnode, right=of r1] (r2) {R2};
\node[smallnode, right=of r2] (r3) {R3};
\node[smallnode, right=of r3] (r4) {R4};
\node[smallnode, right=of r4] (r5) {R5};

\node[container] (container) {};

\node[kernel] at (r1.north) {Read Kernel};
\node[kernel] at (container.north) {Compute Kernel};
\node[kernel] at (r5.north) {Write Kernel};

\node[smallnode, below=of r1] (router0) {Router0};
\node[smallnode, below=of r3] (engine) {Move, Vector, Matrix Engines};
\node[smallnode, below=of r5] (router1) {Router1};

\draw[->] (r1) -- (router0);
\draw[->] (container.south) -- (engine);
\draw[->] (r5) -- (router1);

\node[smallnode, below=of engine] (sram) {SRAM with Buffers};

\draw[->] (router0) -- (sram);
\draw[<->] (engine) -- (sram);
\draw[->] (sram) -- (router1);

\draw[<-] (router0.west) -- ++(-0.4, 0);
\draw[->] (router1.east) -- ++(0.4, 0);
\end{tikzpicture}
\caption{Inside a Tensix core. R1 represents the first RISC-V core. The kernels run on RISC-V cores and the cores control other components. The routers are connected to the NoC and exchange data via Buffers in SRAM with the engines.}
\label{fig:tensix_core}
\end{figure}

\section{Tenstorrent Grayskull e150}
The Tenstorrent Grayskull e150 card supports PCIe 4.0 x16 and is based on a dataflow architecture \cite{dennis1974}. It consists of a \(10 \times 12\) rectangular grid of Tensix cores (See Figure \ref{fig:tensix_core}) connected by a Network-on-Chip \cite{grayskull-noc} (NoC, See Figure \ref{fig:noc_topology}). The cores in the last row cannot be used for computation but still provide memory. By the definition of the dataflow architecture, each Tensix core executes its individual instructions depending on the flow of data through the grid (e.g., the availability of input data), but otherwise, it operates independently of other cores.

Tensix cores operate on tiles of \(32 \times 32\) scalars in various data formats, including Bfloat16. This format has the same number of exponent bits as Float32, but the mantissa is 16 bits shorter. So it has the same range, but reduced precision. However, the precision is sufficient for many operations in machine learning and its use leads to reduced memory consumption and runtime. Therefore, all the following implementations for Grayskull utilize Bfloat16. 

Each tile together with a header containing routing information and a unique id is sent as a packet via the NoC. The NoC is two-dimensional, bi-directional, and has a torus topology (opposite ends are connected). Each Tensix core is connected to the NoC with a bandwidth of \(192\,\text{GB/s}\). 

The Grayskull e150 operates at a maximum clock speed of \(1.2\,\text{GHz}\) and a maximum power of \(200\,\text{W}\). The card has eight channels of LPDDR4 memory at the top and bottom edges of the grid with \(8\,\text{GB}\) total capacity and \(118.4\,\text{GB/s}\) bandwidth (compared to \(80\,\text{GB}\) HBM with \(2\,\text{TB/s}\) of an Nvidia H100 PCIe). It has a computational performance of 92 and 332 TFLOPs for 16- and 8-bit floats respectively (1513 and 3026 TFLOPs for Nvidia H100 PCIe). Keep in mind that the H100 PCIe is approximately \(30\times\) more expensive for the general public. The chip is fabricated using a \(12\,\text{nm}\) process and the area of the core grid is \(477\,\text{mm}^2\). 

\subsection{Tensix core}
A Tensix core has five RISC-V \cite{waterman2014} cores, two routers, a data movement engine, a compute engine and \(1\,\text{MB}\) of SRAM. Since the SRAM has a bandwidth of \(384\,\text{GB/s}\), 120 Tensix cores accessing their memory in parallel would have a bandwidth of approximately 46 TB/s (compared to approximately \(6\,\text{TB/s}\) L2 read bandwidth \cite{lam2023} of an Nvidia H100 PCIe). The routers are connected to the NoC and move tiles in and out of buffers in SRAM. The data movement engine includes a packer and unpacker for moving tiles between SRAM and the compute engine. The compute engine consists of a matrix engine (FPU, not floating-point unit) and vector engine (SFPU). The RISC-V cores dispatch instructions to the other components, which is also called ``driving'' those components.

Incoming tiles are processed in the following way. The first RISC-V core drives router0 to push the incoming tiles from the NoC to a circular buffer in SRAM. The second RISC-V core drives the packer to pop the tiles from the circular buffer to the compute engine. The third core drives the FPU and SFPU to compute mathematical operations. The fourth core drives the unpacker to push the tiles from the compute engine back to another circular buffer in SRAM. And finally, the fifth core drives router1 to pop tiles from the circular buffer back to the NoC.

Because of this highly parallel design, data movement and computation run in parallel which keeps the compute engine busy.

\section{Tenstorrent Software}
For programming Tenstorrent hardware there is the high-level, top-down software stack TT-Buda \cite{tt-buda} and the low-level, bottom-up C++ software stack TT-Metalium \cite{tt-metalium}. 
\subsection{TT-Buda: high-level, top-down}
With TT-Buda, AI models can be defined in external frameworks (e.g., PyTorch \cite{paszke2019}, Tensorflow \cite{abadi2015}) or directly in its Python interface PyBuda. Then, it compiles the model to a binary running on the hardware. The compiler is composed of a frontend and a backend. 

The first step of the frontend is compiling a model from an external framework to a unified intermediate representation using Apache TVM \cite{chen2018}, which is an open-source machine learning compiler framework. The Tenstorrent TVM backend translates this intermediate representation into PyBuda API calls. Executing those calls with dummy tensors creates an execution trace, which is used to construct an initial graph where the nodes represent mathematical operations. These operations are then decomposed into lower-level operations. Optimizations such as constant folding and operation reordering are performed, and the graph is lowered to another intermediate representation. This representation consists of operations based on hardware kernels implemented by the compiler's backend. 
Finally, the balancer assigns resources such as the number of Tensix cores to each operation. The placer then arranges the operations spatially on the grid. The result is a human-readable intermediate representation in YAML format called Netlist, which is passed to the compiler's backend.

The Netlist defines a graph where nodes represent mathematical operations and edges data movement operations. The backend is composed of two independent compile paths. The first one compiles the high level C++ kernels (HLK) of math operations (nodes) to binaries running on Tensix cores. The second path translates data movement operations (edges) into a data movement program and compiles it into another binary running on Tensix cores.

\subsection{TT-Metalium: low-level, bottom-up}
A simple program using TT-Metalium consists of a host program and separate kernels for reading, computing and writing. The host program runs on the CPU and the kernels on Tensix cores. Different Tensix cores can run different kernels. The reader kernel runs on the first RISC-V core of a Tensix core reading tiles from the card's DRAM to SRAM. The compute kernel runs on the three middle RISC-V cores. The writer kernel runs on the last RISC-V core writing tiles from SRAM to the card's DRAM.\footnote{The reader and writer kernels can also be swapped such that the reader runs on the last and the writer on the first RISC-V core.}

A simple host program first instantiates a \textit{Device}, a \textit{Program} and a \textit{CommandQueue} object. The \textit{Device} object represents the Grayskull card. The \textit{Program} object represents the kernels together with input arguments and their spatial placement on the core grid. Once the host program pushes a command to the \textit{CommandQueue}, it is executed on the card. Because the queue allows non-blocking execution of commands, the host program does not have to wait for their completion. 

Sequentially, the host program adds kernels together with their compile-time arguments and spatial Tensix core placement to the \textit{Program} object. Then, it can pass different runtime-arguments such as memory addresses to each Tensix core individually. To store tiles, it allocates buffers in the card's DRAM and circular buffers in the Tensix core's SRAM. Subsequently, data in the host's DRAM is tilized to ensure that elements of each tile are stored consecutively in memory. To write those tiles to the card's DRAM buffer, a \textit{WriteBuffer} command is pushed to the \textit{CommandQueue}. Next, the \textit{Program} object is pushed to the \textit{CommandQueue} to execute the kernels. Finally, a ReadBuffer command is pushed to the \textit{CommandQueue} to read tiles from the card's DRAM back to the host's DRAM and the data gets untilized.

Most of the relevant machine learning operations are already implemented using TT-Metalium and accessible through the Python library TT-NN, which acts as a high-level interface to TT-Metalium.

\section{Multi-Head Self-Attention}
Basic multi-head self-attention \cite{vaswani2017} is defined as:
\begin{align*}
\text{MultiHead}(X) &= \text{Concat}(\text{head}_1, \ldots, \text{head}_h) W^O \\
\text{head}_i &= \text{Attention}(X W_i^Q, X W_i^K, X W_i^V) \\
\text{Attention}(Q, K, V) &= \text{softmax}\left(\frac{Q K^T}{\sqrt{d_k}}\right) V
\end{align*}
where:
\begin{itemize}
    \item \( X \in \mathbb{R}^{n \times d_{\text{model}}} \): Input matrix
    \item \( W_i^Q, W_i^K \in \mathbb{R}^{d_{\text{model}} \times d_k}, W_i^V \in \mathbb{R}^{d_{\text{model}} \times d_v}, \\ W^O \in \mathbb{R}^{h d_v \times d_{\text{model}}} \): Parameter matrices for head \(i\)
    \item \( d_k, d_v \): Key and value dimensions
    \item \( d_{\text{model}} \): Model dimension (of input embeddings)
    \item \( h \): Number of attention heads
    \item \( n \): Batch size (number of tokens in sequence)
\end{itemize}

First, the input vectors are linearly transformed into queries, keys and values by matrix multiplication with the corresponding parameter matrices. Then, the query and key matrices are multiplied to produce a \(n \times n\) matrix of attention scores.  These are scaled by $\frac{1}{\sqrt{d_k}}$ mainly for training stability. The Softmax function is applied to the scaled attention score matrix to calculate the attention weights. By matrix multiplication of the attention weights with the values, effectively a weighted summation of values is calculated for each position in the token sequence. Higher weighted values are prioritized. Hence, the term \textit{attention}. This process is repeated for the other heads with different parameters. Finally, the results are concatenated and linearly transformed to produce the output matrix.

The computation of queries, keys and values, the multiplication of attention weights with those values and the final linear transformation are basic matrix multiplications that were already implemented for the Grayskull architecture by Tenstorrent \cite{mm-implementation}. Concatenation is a trivial operation. Therefore, this work focuses on the efficient computation of \(\text{softmax}\left(\frac{Q K^T}{\sqrt{d_k}}\right)\) for a single head.

\section{Matrix Multiplication on Grayskull}
\label{sec:matrix_multiplication}
To understand matrix multiplication on Grayskull \cite{mm-implementation}, imagine that the first input matrix flows from the left to the right of the Tensix core grid, while simultaneously the second input matrix flows from the top to the bottom. In the end, the output matrix is laid out on the Tensix core grid such that the top left element of the output matrix is stored on the top left Tensix core and the bottom right element on the bottom right Tensix core. The outputs stored on each core were computed exclusively on the same core by matrix multiplication. 

Concretely, it is implemented in the following way. First, some cores have to read tiles from the card's DRAM into their SRAM. The first column of the Tensix core grid reads a specified number of columns from the first input matrix and the first row of the core grid the same number of rows from the second input matrix. Each core reads a block of tiles. The block width/height of the first/second input matrix is the specified number. Then, each core in the first column multicasts its block to all cores in its row and each core in the first row to all cores in its column. This process repeats with the next blocks until all tiles of the input matrices are read. Given input matrices \(A\) with dimensions \(8 \times 4\), \(B\) with \(4 \times 8\), a \(2 \times 2\) core grid and a specified block width/height of 2, the second core of the first column would read and multicast the two bottom \(4 \times 2\) blocks of A.

Simultaneously, all cores compute matrix multiplication with each incoming block from the first input matrix and the corresponding block from the second input matrix. They add each result to an intermediate output block. Once all blocks are processed, this intermediate output block is the final output block and all cores write their output blocks back to the card's DRAM.

\begin{table*}[t]
\centering
\small
\begin{tabular}{|c|c|c|c|c|c|c|c|c|}
\hline
\textbf{n} & \multicolumn{2}{c|}{\textbf{CPU}} & \multicolumn{6}{c|}{\textbf{Grayskull}} \\
\cline{2-9}
of \(n \times n\) & \multicolumn{2}{c|}{\textbf{Softmax}} & \multicolumn{3}{c|}{\textbf{Softmax}} & \multicolumn{3}{c|}{\textbf{Matrix Multiply + Scaling + Softmax}} \\
\cline{2-9}
& \textbf{Recomputing} & \textbf{Caching} & \textbf{Single-Core} & \multicolumn{2}{c|}{\textbf{Multi-Core}} & \textbf{Total} & \multicolumn{2}{c|}{\textbf{Kernel}} \\
\cline{4-6} \cline{8-9}
& & &{\textbf{Total}} & \textbf{Total} & \textbf{Kernel} & & \textbf{Total} & \textbf{Softmax} \\
\hline
1024 & 7.37 & 1.15 & 2.67 & 0.261 & 0.178 & 0.29 & 0.174 & 0.0644 \\
\hline
2048 & 28.8 & 4.51 & 10.9 & 0.582 & 0.524 & 0.686 & 0.573 & 0.26 \\
\hline
4096 & 115 & 17.9 & 43.6 & 1.90 & 1.83 & 2.14 & 2 & 1.05 \\
\hline
8192 & 460 & 73.1 & 177 & 7.26 & 7.04 & {---} & {---} & {---} \\
\hline
16384 & 1850 & 297 & {---} & {---} & {---} & {---} & {---} & {---} \\
\hline
\end{tabular}
\caption{Experimental results for the runtime of different implementations in milliseconds. The total runtime of Grayskull implementations is measured on the host and includes overhead like dispatching the kernel. The kernel runtime without overhead is measured on Grayskull. The total CPU and Grayskull runtimes were averaged over 100 iterations and the kernel runtimes over 10. If a runtime is not displayed, the input matrix is too large.}
\label{tab:implementation_runtimes}
\end{table*}

\section{Softmax}
\label{sec:softmax}
The Softmax function applied to a matrix is:
\[
\sigma(Z)_{ij} = \frac{e^{Z_{ij}}}{\sum_{k=1}^n e^{Z_{ik}}}
\]
with:
\begin{itemize}
  \item \(Z_{ij}\): Element in the \(i\)-th row and \(j\)-th column of the \(m \times n\) input matrix.
  \item \(\sigma(Z)_{ij}\): Element in the \(i\)-th row and \(j\)-th column of the \(m \times n\) output matrix.
\end{itemize}

The natural exponential function is applied to all elements of \(Z\) and each \(e^{Z_{ij}}\) is normalized by the sum of all elements in its row \(i\). Therefore, each output element \(\sigma(Z)_{ij}\) depends on all \(e^{Z_{i1}}, \ldots, e^{Z_{in}}\) of the same row \(i\) but not on exponentials of other rows. So all \(e^{Z_{i1}}, \ldots, e^{Z_{in}}\) have to be computed before \(\sigma(Z)_{ij}\). Caching them avoids the need to recompute \(e^{Z_{ij}}\) for each \(\sigma(Z)_{ij}\). However, the capacity of the cache must be sufficient to store an entire row. 

When the input elements \(Z_{ij}\) are large, computing \(e^{Z_{ij}}\) and \(\sum_{k=1}^n e^{Z_{ik}}\) leads to a risk of numerical overflow. To reduce it, this equation can be used:
\begin{equation}
\frac{e^{Z_{ij} - m_i}}{\sum_{k=1}^{K} e^{Z_{ik} - m_i}} = \frac{e^{Z_{ij}} / e^{m_i}}{\sum_{k=1}^{K} e^{Z_{ik}} / e^{m_i}} = \frac{e^{Z_{ij}}}{\sum_{k=1}^{K} e^{Z_{ik}}}
\end{equation}
where $m_i = \max_j Z_{ij}$.

From each input element the maximum of its row is subtracted. Therefore, the maximum input element is \(0\), which reduces the risk of numerical overflow.

\subsection{Softmax on CPU}
The CPU implementation of the Softmax function serves as a baseline, and to observe the effect of caching the exponentials. It processes the input matrix row by row. First, it searches the maximum and subtracts it from all elements in the current row. Then, it exponentiates the elements and sums them. Finally, it normalizes the exponentials by the sum.

To measure the effect of caching, one implementation caches the exponentials in DRAM and SRAM at the summing step, and another one recomputes them at the final step. Both are written in C and use 32-bit floats. Since the attention score matrix is \(n \times n\) where \(n\) is the input token sequence length, this work focuses mainly on square matrices.

\subsubsection{\textbf{Maximum sequence length}}
Because this CPU implementation is built for experimental evaluation, it allocates memory for both input and output data to avoid resetting the inputs at each profiling iteration, which would distort the results.
However, A CPU implementation for production could operate \textit{in-place}, meaning that it operates directly on the input data. The maximum dimension for such an implementation would be \(n_{\text{max}} = \sqrt{\frac{\text{memory}}{4}}\), because a float is 4 bytes. For 8 GiB of DRAM that is \(n_{\text{max}} = 46340\) and means we could process a sequence of \(46340\) input tokens. 
\subsubsection{\textbf{Time and memory complexity}}
Each element is accessed and processed three times with constant time complexity \(O(1)\). Therefore, the total time complexity is \(\Theta(n^2)\), which can be observed in the experimental results (see Table~\ref{tab:implementation_runtimes}). Clearly, an in-place implementation has a memory complexity of \(\Theta(n^2)\). Since the other implementation allocates twice as much memory, it has also \(\Theta(n^2)\).

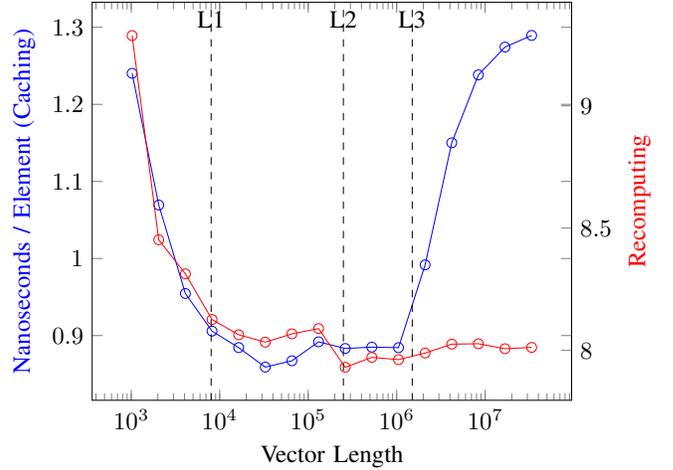
\begin{figure}[t]
\centering
\begin{tikzpicture}[scale=0.93]
\begin{axis}[
    xlabel={Vector Length},
    xmode=log,
    log basis x=10,
    ylabel={Nanoseconds / Element (Caching)},
    ylabel style={blue},
]
\addplot[
    color=blue,
    mark=o,
    ]
    table [col sep=comma, x=length, y=cache] {data/cpu.csv};
\end{axis}
\begin{axis}[
    axis x line=none,
    xmode=log,
    log basis x=10,
    axis y line*=right,
    ylabel={Recomputing},
    ylabel style={red},
]
\addplot[
    color=red,
    mark=o,
    ]
    table [col sep=comma, x=length, y=recompute] {data/cpu.csv};
\draw[dashed] (axis cs:8 * 1e3,\pgfkeysvalueof{/pgfplots/ymin}) -- (axis cs:8 * 1e3,\pgfkeysvalueof{/pgfplots/ymax});
\node[below] at (axis cs:8 * 1e3,\pgfkeysvalueof{/pgfplots/ymax}) {L1};
\draw[dashed] (axis cs:2.5 * 1e5,\pgfkeysvalueof{/pgfplots/ymin}) -- (axis cs:2.5 * 1e5,\pgfkeysvalueof{/pgfplots/ymax});
\node[below] at (axis cs:2.5 * 1e5,\pgfkeysvalueof{/pgfplots/ymax}) {L2};
\draw[dashed] (axis cs:1.5 * 1e6,\pgfkeysvalueof{/pgfplots/ymin}) -- (axis cs:1.5 * 1e6,\pgfkeysvalueof{/pgfplots/ymax});
\node[below] at (axis cs:1.5 * 1e6,\pgfkeysvalueof{/pgfplots/ymax}) {L3};
\end{axis}
\end{tikzpicture}
\caption{Effect of caching the exponentials. Note, that the two y-axes have different scales. All CPU experiments were conducted on a single core of an Intel i5-6500 processor with 8 GB DDR4 memory, running Ubuntu 20.04. It has 32 KB L1 cache per core, 1 MB shared L2 and 6 MB shared L3 cache.}
\label{fig:softmax_cpu_caching}
\end{figure}

\subsubsection{\textbf{Effect of caching}}
The experimental results also show that the implementation caching the exponentials is approximately \(6\times\) faster than the one recomputing them on this specific computer configuration. 

Figure~\ref{fig:softmax_cpu_caching} shows the runtime per element across varying length of a vector for both the caching and the recomputing implementation. Note, that their y-axes have different scales. Up to a vector length of \(32768\), the values of both implementations decrease. A likely reason is that there is a constant overhead and its relative effect on the total runtime decreases for larger vectors. Starting from a vector length of \( 2^{21} \approx 2.09 \times 10^6 \), the runtime per element of the caching implementation increases steeply until it plateaus, while the values of the recomputing implementation remain approximately constant. The reason is that the L3 cache of this specific CPU has a capacity of \(6 \text{ MB} = 1.5 \times 10^6 \text{ Floats}\), which is smaller than the number of floats in the vector. With the caching implementation, this leads to cache misses. As a result, the CPU has to fetch the exponentials from DRAM, which is significantly slower. For the L1 and L2 cache the same effect cannot be observed, likely due to less variation in bandwidth among the caches compared to the significant variation between bandwidth of L3 cache and DRAM. 

In conclusion, the caching implementation is significantly faster than the recomputing one for all sequence length \(n\), because it utilizes the DRAM for caching. But beginning from \(n = 1.5 \times 10^6\), the runtime per element increases, because it cannot utilize the CPU caches anymore.

\begin{figure}[t]
    \centering
    \begin{tikzpicture}
        \begin{axis}[
            bar width=20pt,
            xlabel={Row in Tensix core grid},
            xtick=data,
            xticklabels={1, 5, 9},
            enlarge x limits=0.5,
            ybar stacked,
            ymin=0,
            ylabel={Runtime in milliseconds},
            legend style={at={(0.5,-0.2)}, anchor=north, legend columns=4},
        ]
        \addplot table[x=core_row, y=load1,col sep=comma] {data/grayskull/softmax_distribution.csv};
        \addplot table[x=core_row, y=max1,col sep=comma] {data/grayskull/softmax_distribution.csv};
        \addplot table[x=core_row, y=subtract1,col sep=comma] {data/grayskull/softmax_distribution.csv};
        \addplot table[x=core_row, y=exponentiate1,col sep=comma] {data/grayskull/softmax_distribution.csv};
        \addplot table[x=core_row, y=sum1,col sep=comma] {data/grayskull/softmax_distribution.csv};
        \addplot table[x=core_row, y=normalize1,col sep=comma] {data/grayskull/softmax_distribution.csv};
        \addplot table[x=core_row, y=load2,col sep=comma] {data/grayskull/softmax_distribution.csv};
        \addplot table[x=core_row, y=max2,col sep=comma] {data/grayskull/softmax_distribution.csv};
        \addplot table[x=core_row, y=subtract2,col sep=comma] {data/grayskull/softmax_distribution.csv};
        \addplot table[x=core_row, y=exponentiate2,col sep=comma] {data/grayskull/softmax_distribution.csv};
        \addplot table[x=core_row, y=sum2,col sep=comma] {data/grayskull/softmax_distribution.csv};
        \addplot table[x=core_row, y=normalize2,col sep=comma,col sep=comma] {data/grayskull/softmax_distribution.csv};
        \addplot table[x=core_row, y=load3,col sep=comma] {data/grayskull/softmax_distribution.csv};
        \addplot table[x=core_row, y=max3,col sep=comma] {data/grayskull/softmax_distribution.csv};
        \addplot table[x=core_row, y=subtract3,col sep=comma] {data/grayskull/softmax_distribution.csv};
        \addplot table[x=core_row, y=exponentiate3,col sep=comma] {data/grayskull/softmax_distribution.csv};
        \addplot table[x=core_row, y=sum3, col sep=comma] {data/grayskull/softmax_distribution.csv};
        \addplot table[x=core_row, y=normalize3,col sep=comma] {data/grayskull/softmax_distribution.csv}; \legend{load1,max1,subtract1,exponentiate1,sum1,normalize1,load2,max2,subtract2,exponentiate2,sum2,normalize2,load3,max3,subtract3,exponentiate3,sum3,normalize3}
        \end{axis}
    \end{tikzpicture}
    \caption{Runtime distribution of the Softmax kernel with \(8192 \times 8192\) input matrix. Measured on the compute core (3rd of 5 RISC-V cores) of the first Tensix core from three different rows in the core grid. For example, load1 means loading the first row of tiles. The Tensix core in the first row processes one additional row of tiles.}
    \label{fig:softmax_grayskull_runtime_distribution}
\end{figure}
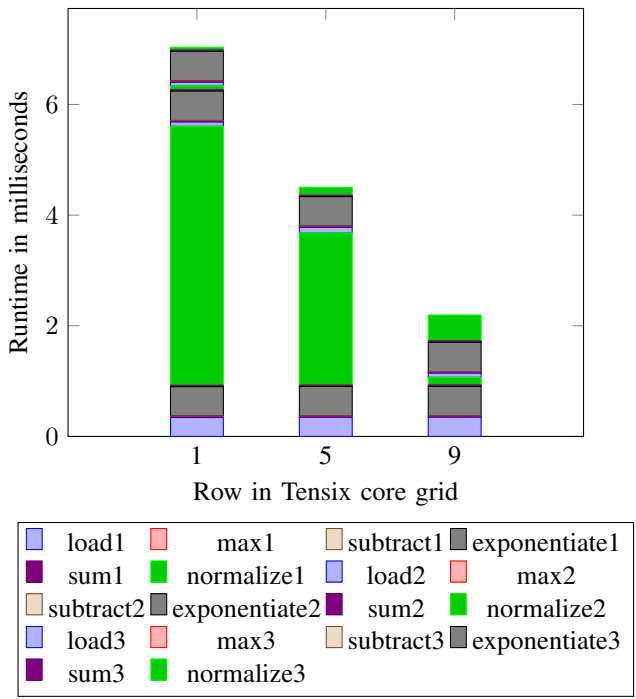

\begin{figure*}[t]
\centering
\begin{tikzpicture}
\begin{axis}[
    width=\textwidth,
    height=0.5\textwidth,
    xlabel={$n$ (of $n \times n$ input matrix)},
    xtick={1024, 2048, 4096},
    ylabel={Runtime in microseconds},
    legend pos=north west,
    stack plots=y,
    area style,
]
\legend{Matrix Multiplication, Attention Score Scaling, Reduce Max, Reduce Max other Cores, Subtract Max, Exponentiate, Reduce Sum, Reduce Sum other Cores, Normalize with Sum}
\addplot+[fill, blue] table [x=dimension, y=matrix_multiply, col sep=comma] {data/grayskull/fused.csv} \closedcycle;
\addplot+[fill, red] table [x=dimension, y=attention_scaling, col sep=comma] {data/grayskull/fused.csv} \closedcycle;
\addplot+[fill, green] table [x=dimension, y=reduce_max, col sep=comma] {data/grayskull/fused.csv} \closedcycle;
\addplot+[fill, orange] table [x=dimension, y=reduce_max_other_cores, col sep=comma] {data/grayskull/fused.csv} \closedcycle;
\addplot+[fill, purple] table [x=dimension, y=subtract_max, col sep=comma] {data/grayskull/fused.csv} \closedcycle;
\addplot+[fill, cyan] table [x=dimension, y=exponentiate, col sep=comma] {data/grayskull/fused.csv} \closedcycle;
\addplot+[fill, brown] table [x=dimension, y=reduce_sum, col sep=comma] {data/grayskull/fused.csv} \closedcycle;
\addplot+[fill, lime] table [x=dimension, y=reduce_sum_other_cores, col sep=comma] {data/grayskull/fused.csv} \closedcycle;
\addplot+[fill, magenta] table [x=dimension, y=normalize, col sep=comma] {data/grayskull/fused.csv} \closedcycle;
\end{axis}
\end{tikzpicture}
\caption{Runtime distribution of the fused kernel across varying input dimensions. Measured on the compute core (3rd of 5 RISC-V cores) of the fastest Tensix core.}
\label{fig:run_dist}
\end{figure*}
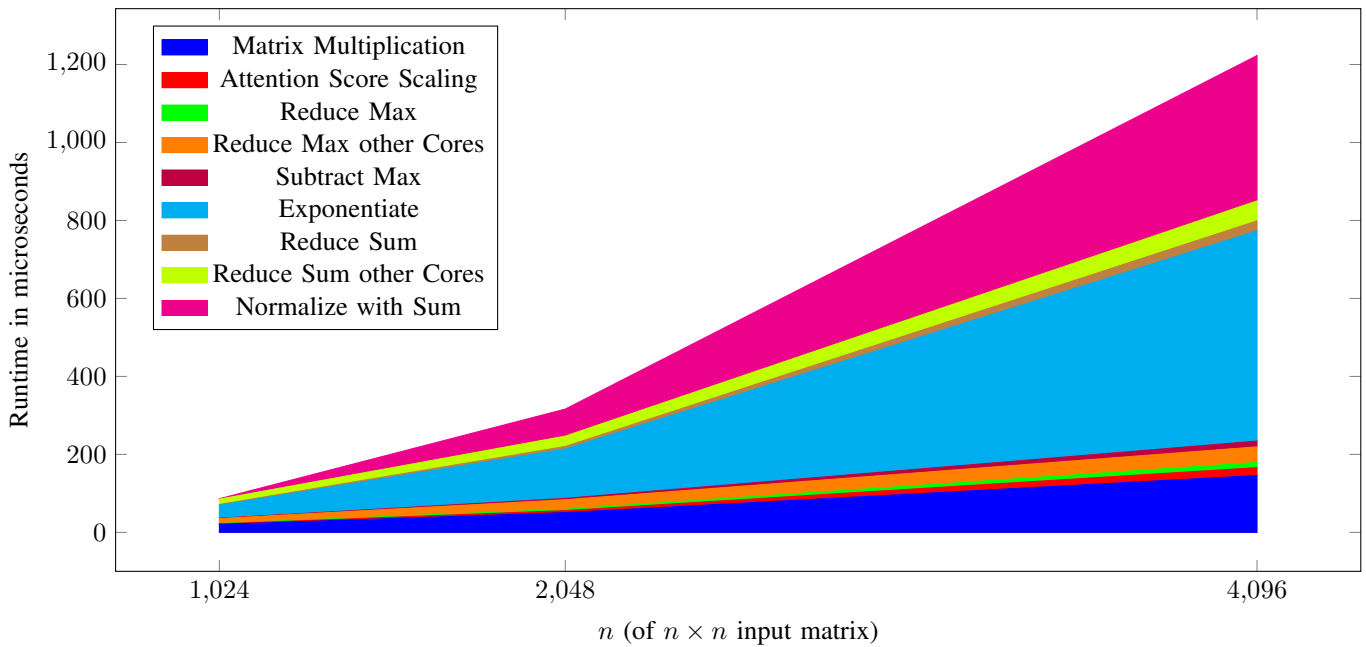

\subsection{Softmax on Grayskull}
The Softmax implementation for the Grayskull architecture can utilize a single or multiple Tensix cores. Because a row of the output matrix only depends on the same row and not on other rows of the input matrix, different rows can be processed on different cores in parallel. Since the cores compute on \(32 \times 32\) tiles, the matrices are partitioned into tiles, and rows consist of tiles instead of scalars. To determine the number of rows each core processes, division with remainder is performed with the total number of rows and the number of cores:
\[
\frac{n\_rows}{n\_cores} = min\_rows\_per\_core\text{, rest: }n\_cores\_plus\_one
\]

The quotient determines the minimum number of rows that each core processes. Additionally, different remaining rows are distributed to different cores starting from the upper left core, in a row-wise manner. Each Tensix core processes its rows one at a time and performs the same computation as the CPU implementation. However, it is only responsible for a subset of the rows, computes more efficiently on tiles and accesses its SRAM explicitly in contrast to implicitly accessing transparent caches.

\subsubsection{\textbf{Maximum sequence length}}
Because the SRAM of a Tensix core has a capacity of \(1\,\text{MB} \approx 488\,\text{Tiles (Bfloat16)}\) and a tile has a width of \(32\), the maximum length of a row is \(488 \times 32 = 15616\,\text{Floats}\). In the context of the attention operation, the maximum sequence length would be \(n_{\text{max}} = 15616\). There is no limit on the number of rows.

\subsubsection{\textbf{Time and memory complexity}}
In contrast to the CPU implementation, it can process up to a constant number of rows in parallel. Since this speedup is just a constant factor, the time complexity of this implementation is \(\Theta(n^2)\) as well. Because each Tensix core has to store an entire row, the memory complexity for the SRAM is \(\Theta(n)\). However, the implementation has to allocate DRAM for the input and output matrices. Hence, the memory complexity is \(\Theta(n^2)\).

\subsubsection{\textbf{Experimental results}}
The number of cores, that process at least one row of tiles, increases from \(n = 1024\) to \(n = 4096\) where all of them do. At \(n = 4096\) the first \(20\) cores process one additional row and at \(n = 8192\) all of them process \(2\) rows while the first \(40\) process one additional row. 

Doubling \(n\) should quadruple the runtime according to the time complexity. However, only the runtime of \(n=8192\) is almost \(4\times\) larger than the one of \(n=4096\). This is because the number of running cores increases from \(n = 1024\) to \(n = 4096\). For the same reason, the Grayskull multi-core vs. single-core speedup increases from approximately \(10\times\) at \(n = 1024\) to \(23\times\) beginning at \(n = 4096\). The Grayskull multi-core vs. CPU caching speedup increases from approximately \(4.4\times\) at \(n = 1024\) to \(10\times\) at \(n = 8192\).

Particularly for sequences with at least \(n = 4096\) input tokens, running the Softmax operation on Grayskull would be significantly faster than on a CPU.

\subsubsection{\textbf{Runtime distribution}}
The distribution of the total runtime across all operations is shown in Figure~\ref{fig:softmax_grayskull_runtime_distribution} for three different Tensix cores.

First, the compute core (third RISC-V core inside a Tensix core) has to wait for the first two RISC-V cores to load the first row into SRAM and its tiles from there to registers. Because the five RISC-V cores work in parallel, the first two cores can already load the next row, while the compute core is processing the current row, ensuring that the compute core does not need to wait for subsequent rows. Hence, the runtime of the first loading operation is significant, while the runtime of the following loading operations are negligible. 

The runtime of the exponentiations is significant and remains constant for different rows and Tensix cores. 

The runtime of the actual normalization is a small constant. However, there is an extreme variance in the runtime of the first normalization operation across Tensix cores in different rows. The likely reason is as follows. Since normalization is the last operation, it sends the results tile by tile to the last two RISC-V cores and each time has to wait for them to write the tile back to DRAM. Due to the topology of the NoC, each Tensix core has to wait for the one below it to write the results to DRAM. Therefore, the operation takes longer in higher rows of the core grid.

The runtime of the reduction operations (max, sum) and of the subtraction operation are negligible.

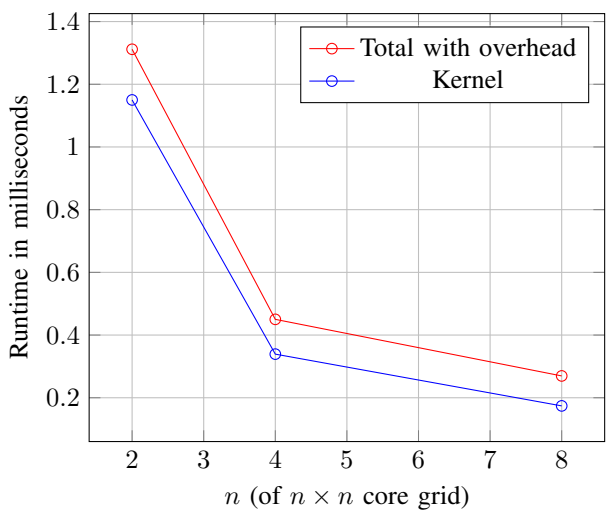
\begin{figure}[b]
    \centering
    \begin{tikzpicture}
        \begin{axis}[
            xlabel={$n$ (of $n \times n$ core grid)},
            ylabel={Runtime in milliseconds},
            legend pos=north east,
            grid=both,
        ]
        \addplot[
            color=red,
            mark=o,
            ]
            table[x=dimension,y=total,col sep=comma] {data/grayskull/softmax_cores.csv};
        \addplot[
            color=blue,
            mark=o,
            ]
            table[x=dimension,y=kernel,col sep=comma] {data/grayskull/softmax_cores.csv};
        \legend{Total with overhead, Kernel}
        \end{axis}
    \end{tikzpicture}
    \caption{Runtime of the fused kernel across varying dimensions of a square grid of cores. Measured on the host with overhead and on Grayskull without overhead.}
    \label{fig:recompute_cache}
\end{figure}

\section{Fused Matrix Multiply + Scaling + Softmax on Grayskull}
The computation of \(\text{softmax}\left(\frac{Q K^T}{\sqrt{d_k}}\right)\) is fundamental to the attention mechanism. However, separate kernels for the matrix multiplication of queries and keys, the scaling of the attention scores, and the application of the Softmax function to those scores have to read intermediate results of the previous kernel from DRAM and write their intermediate results back to it. But DRAM has lower bandwidth, higher latency, and higher energy consumption than SRAM. Additionally, the dispatching of each kernel to Grayskull introduces overhead. This motivates the use of a fused kernel reducing overhead of accessing DRAM and dispatching kernels.

The implementation of the fused kernel builds on top of a multi-core matrix multiplication implementation by Tenstorrent \cite{mm-implementation}(described in Section~\ref{sec:matrix_multiplication}). After the matrix multiplication is performed, the attention score matrix is laid out on the Tensix core grid such that the top left element is stored on the top left Tensix core and the bottom right element on the bottom right core. Each Tensix core stores a subset of the attention score matrix and computes the scaling and Softmax operations on them. Since the scaling operation is defined as element-wise multiplication with \(\frac{1}{\sqrt{d_k}}\), its implementation is trivial. For the Softmax operation, each Tensix core first computes the local maxima of partial rows of scaled attention scores stored in its memory. But complete rows span an entire row of Tensix cores. Therefore, each Tensix core reads the local maxima from all other cores in its row to compute the global maxima of complete rows. Then, it subtracts the maxima from its scaled attention scores and exponentiates them. In the same way, it computes the local sums, then the global sums and normalizes its exponentials with those sums. Finally, each core writes its attention weights to its individual address inside the resulting attention weight matrix in DRAM.

The following focuses on the use of the kernel in the context of the attention operation with the \(n \times d_k\) input matrices \(Q\) and \(K\), as well as a \(n \times n\) output matrix of attention weights. All experiments were conducted with \(d_k = 128\), since this value was used in the popular Llama3\cite{meta2024}. 

\subsubsection{\textbf{Maximum sequence length}}
Because the SRAM of a Tensix core has a capacity of \(5 \times 10^5\) floats in Bfloat16, the largest square matrix it can store is \(707 \times 707\). Since the largest (for the computation utilizable) Tensix core grid is \(9 \times 12\), the square output matrix has \(n_\text{max} = 707 \times 9 = 6363\). However, in practice it is slightly less due to buffers for caching and dataflow. The largest successfully tested power of two was \(n = 4096 \). So with this implementation the maximum number of input tokens in a sequence is approximately \(4096\), which is significantly less than \(n_\text{max} = 15616\) of the dedicated Softmax kernel.

\subsubsection{\textbf{Time and memory complexity}}
Because \(d_k\) is a constant, the dot product to produce a single attention score is \(O(1)\). Since the attention score matrix has dimensions \(n \times n\), the time complexity of the matrix multiplication is \(\Theta(n^2)\). For the same reason, the scaling of attention scores is \(\Theta(n^2) \). Each Tensix core computes the regular Softmax operation. Additionally, it reads the local maxima and sums from other cores in its row, but this is just a constant overhead. Multiple cores provide a significant speedup, but this is just a constant factor. Therefore, the Softmax part is \(\Theta(n^2)\) as well.   
Hence, the time complexity of the fused kernel is \(\Theta(n^2)\), which is also show in the experimental results in Table~\ref{tab:implementation_runtimes}. 

Since the input and output matrices are \(n \times n\), the memory complexity is \(\Theta(n^2)\).

\subsubsection{\textbf{Experimental results}}
Table~\ref{tab:implementation_runtimes} shows, that the speedup of the Softmax operation inside the fused kernel, compared to the dedicated multi-core Softmax kernel, is \(4.1\times\) at \(n=1024\) and decreases to \(1.8\times\) at \(n=4096\). The speedup is mainly due to avoiding DRAM accesses. It decreases, since the number of active cores for the dedicated Softmax kernel increases, while the fused kernel is computed on \(8 \times 8\) Tensix cores for all \(n\).

Because of this speedup and the negligible runtime of the scaling operation, the runtime of the fused kernel is at \(n=1024\) smaller and otherwise only slightly larger than the runtime of the kernel for only the Softmax operation. However, the dispatching overhead of the fused kernel is approximately twice as large and therefore, the total runtime is slightly larger.

Compared to the CPU implementation with caching, the speedup is approximately \(17\times\) for \(n \in \{1024, 2048, 4096\}\).

\subsubsection{\textbf{Runtime distribution}}
The primary factors influencing the runtime are matrix multiplication, exponentiation, and normalization (See Figure~\ref{fig:run_dist}). At \(n = 4096\) normalization takes approximately \(2.6 \times\) longer than matrix multiplication and exponentiation \(1.4 \times\) longer than normalization. So the runtime of the Softmax operation is significantly larger than the one of matrix multiplication. The reduction operations computing the global maxima and sums by reading the local ones from other cores influence the total runtime slightly. The runtime of all other operations is negligible.

Figure~\ref{fig:run_dist} also shows the \(\Theta(n^2)\) time complexity.

\section{Limitations and future Directions}
Currently, the fused kernel computes the attention weights from the queries and keys, but not the queries and keys itself from the input matrix and also not the output matrix with the attention weights, output weights, and values. Future work could try to incorporate those remaining matrix multiplications into the fused kernel to reduce overhead from kernel dispatching and DRAM accesses.

The Softmax implementation inside the fused kernel computes the same global sums and maxima redundantly on all cores. Since all cores run in parallel, this has no negative effect on the runtime. Future work could implement other variants such as computing the global sums and maxima only on one core in each row and broadcasting them to the other cores. Then, one could compare energy, runtime and memory consumption.

Finally, it would be interesting to port the implementation to newer generations (e.g., Tenstorrent Wormhole) and to scale it on multiple cards.

\bibliographystyle{plain}
\bibliography{references}
\end{document}